\documentclass{article}

\usepackage[final,nonatbib]{neurips_2023}

\usepackage[utf8]{inputenc} % allow utf-8 input
\usepackage[T1]{fontenc}    % use 8-bit T1 fonts
\usepackage{hyperref}       % hyperlinks
\usepackage{url}            % simple URL typesetting
\usepackage{booktabs}       % professional-quality tables
\usepackage{amsfonts}       % blackboard math symbols
\usepackage{nicefrac}       % compact symbols for 1/2, etc.
\usepackage{microtype}      % microtypography
\usepackage{xcolor}         % colors

\usepackage{comment}
\usepackage{amsmath}
\usepackage{amsthm}

\DeclareMathOperator*{\argmin}{arg\,min}
\usepackage{algorithm}
\usepackage{algorithmic,esint}
\usepackage{enumitem}
\usepackage{graphicx}
\usepackage{newfloat}
\usepackage{listings}
\usepackage{subfig}
\usepackage{adjustbox}
\usepackage{wrapfig}
\newtheorem{theorem}{Theorem}

\newtheorem{proposition}[theorem]{Proposition}

\newtheorem{assumption}{Assumption}

\newcommand{\model}{\textbf{SILD} }
\newcommand{\modelnosp}{\textbf{SILD}}
\newcommand{\red}[1]{\textcolor{black}{#1}}
\newcommand{\ms}[2]{{#1}\scriptsize{$\pm$#2}}
\newcommand{\msone}[2]{\bf {#1}\scriptsize{$\pm$#2}}
\newcommand{\mstwo}[2]{\underline{{#1}\scriptsize{$\pm$#2}}}

\title{Spectral Invariant Learning for Dynamic Graphs under Distribution Shifts}

\author{%
Zeyang Zhang\textsuperscript{1}\thanks{This work was done during the author's internship at Alibaba Group},\;
Xin Wang\textsuperscript{1}\thanks{Corresponding authors},\;
Ziwei Zhang\textsuperscript{1},\;
Zhou Qin\textsuperscript{2},\\
\textbf{Weigao Wen}\textsuperscript{2},\;
\textbf{Hui Xue}\textsuperscript{2},\;
\textbf{Haoyang Li}\textsuperscript{1},\;
\textbf{Wenwu Zhu}\textsuperscript{1}\footnotemark[2]\\
\textsuperscript{1}Department of Computer Science and Technology, BNRist, Tsinghua University,\;
\textsuperscript{2}Alibaba Group\\
\small \texttt{zy-zhang20@mails.tsinghua.edu.cn},\;
\small \texttt{\{xin\_wang, zwzhang\}@tsinghua.edu.cn},\\
\small \texttt{\{qinzhou.qinzhou, weigao.wen, hui.xueh\}@alibaba-inc.com},\\
\small \texttt{lihy18@mails.tsinghua.edu.cn},\;
\small \texttt{wwzhu@tsinghua.edu.cn}\\
}

\begin{document}

\maketitle

\begin{abstract}

Dynamic graph neural networks (DyGNNs) currently struggle with handling distribution shifts that are inherent in dynamic graphs.
Existing work on DyGNNs with out-of-distribution settings only focuses on the time domain, failing to handle cases involving distribution shifts in the spectral domain. 
In this paper, we discover that there exist cases with distribution shifts unobservable in the time domain while observable in the spectral domain, 
and propose to study distribution shifts on dynamic graphs in the spectral domain for the first time.
However, this investigation poses two key challenges: 
i) it is non-trivial to capture different graph patterns that are driven by various frequency components entangled in the spectral domain; and 
ii) it remains unclear how to handle distribution shifts with the discovered spectral patterns. 
To address these challenges, we propose Spectral Invariant Learning for Dynamic Graphs under Distribution Shifts (\modelnosp), which can handle distribution shifts on dynamic graphs by capturing and utilizing invariant and variant spectral patterns. Specifically, we first design a DyGNN with Fourier transform to obtain the ego-graph trajectory spectrums, allowing the mixed dynamic graph patterns to be transformed into separate frequency components. We then develop a disentangled spectrum mask to filter graph dynamics from various frequency components and discover the invariant and variant spectral patterns. Finally, we propose invariant spectral filtering, which encourages the model to rely on invariant patterns for generalization under distribution shifts. Experimental results on synthetic and real-world dynamic graph datasets demonstrate the superiority of our method for both node classification and link prediction tasks under distribution shifts. 
\end{abstract}

\section{Introduction}
Dynamic graph neural networks (DyGNNs) have achieved remarkable success in many predictive tasks over dynamic graphs~\cite{skarding2021foundations,zhu2022learnable}.
Existing DyGNNs exhibit limitations in handling distribution shifts, which naturally exist in dynamic graphs due to multiple uncontrollable factors 
~\cite{brown1992survivorship,berk1983introduction,zhu2021shift,kim2021reversible}.  
Existing work on out-of-distribution generalized DyGNNs focuses on handling distribution shifts in the time domain. For example, DIDA~\cite{zhang2022dynamic} utilizes dynamic graph attention to mask the graph trajectories to capture the invariant patterns on dynamic graphs, which assumes that in the time domain, the distribution shift is observable and the invariant and variant patterns can be easily disentangled. 

\begin{figure}
\centering
\includegraphics[width=1\textwidth]{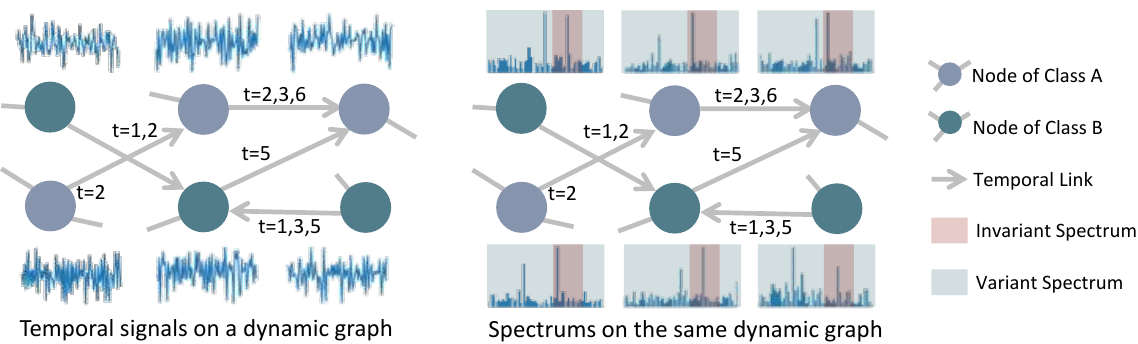}    
\caption{ An illustration example: the graph dynamics from different frequency components are entangled in the temporal domain, while it is much easier to distinguish different frequency components by masking the spectrums in the spectral domain. In this case, the frequency components in the invariant spectrums determine the node labels, while the relationship between the variant spectrums and labels is not stable under distribution shifts. }
\label{fig:example}
\end{figure}

However, there exist cases that the distribution shift is unobservable in the time domain while observable in the spectral domain, as shown in Figure~\ref{fig:example}. The shift in frequency components can be clearly observed in the spectral domain, while these components are indistinguishable in the time domain. Moreover, in real-world applications, the observed dynamic graphs usually consist of multiple mixed graph structural and featural dynamics from various frequency components~\cite{mondal2012managing,dang2016timearcs,barunik2020dynamic}. 

To address this problem, in this paper, we study the problem of handling distribution shifts on dynamic graphs in the spectral domain for the first time, which poses the following two key challenges:
i) it is non-trivial to capture different graph patterns that are driven by various frequency components entangled in the spectral domain, and 
ii) it remains unclear how to handle distribution shifts with the discovered spectral patterns.

To tackle these challenges, we propose \underline Spectral \underline Invariant \underline Learning for \underline Dynamic Graphs under Distribution Shifts  (\modelnosp\footnote{The codes are available at https://github.com/wondergo2017/sild.}). 
Our proposed \model model can effectively handle distribution shifts on dynamic graphs by discovering and utilizing invariant and variant spectral patterns.
Specifically, we first design a DyGNN with Fourier transform to obtain the ego-graph trajectory spectrums so that the mixed graph dynamics can be transformed 
into separate frequency components. Then we develop a disentangled spectrum mask that leverages the amplitude and phase information of the ego-graph trajectory spectrums to obtain invariant and variant spectrum masks so that graph dynamics from various frequency components can be filtered. Finally, we propose an invariant spectral filtering that discovers the invariant and variant patterns via the disentangled spectrum masks, and minimize the variance of predictions with exposure to various variant patterns. 
As such, \model is able to exploit invariant patterns to make predictions under distribution shifts.
Experimental results on several synthetic and real-world datasets, including both node classification and link prediction tasks, demonstrate the superior performance of our \model model compared to state-of-the-art baselines under distribution shifts. 
To summarize, we make the following contributions:

\begin{itemize}[leftmargin=0.5cm]
    \item We propose to study distribution shifts on dynamic graphs in the spectral domain, to the best of our knowledge, for the first time.
    
    \item We propose \underline Spectral \underline Invariant \underline Learning for \underline Dynamic Graphs under Distribution Shifts  (\modelnosp), which can handle distribution shifts on dynamic graphs in the spectral domain.
    
    \item We employ DyGNN with Fourier transform to obtain the node spectrums, design a disentangled spectrum mask to obtain invariant and variant spectrum masks in the spectral domain, and propose the invariant spectral filtering mechanism so that \model is able to handle distribution shifts.
    
    \item We conduct extensive experiments on several synthetic and real-world datasets, including both node classification and link prediction tasks, to demonstrate the superior performance of our method compared to state-of-the-art baselines under distribution shifts.
    
\end{itemize}

\section{Problem Formulation and Notations}
\paragraph{Dynamic Graphs} A dynamic graph can be represented as $\mathcal{G}=(\{\mathcal{G}^{t}\}_{t=1}^{T})$, where $T$ represents the total number of time stamps, and each $\mathcal{G}^t=(\mathcal{V}^t,\mathcal{E}^t)$ corresponds to a graph snapshot at time stamp $t$ with the node set $\mathcal{V}^t$ and the edge set $\mathcal{E}^t$ . For simplicity, we also represent a graph snapshot as $\mathcal{G}^t=(\mathbf{X}^t,\mathbf{A}^t)$, which includes the node feature matrix $\mathbf{X}^t$ and the adjacency matrix $\mathbf{A}^t$. We further denote a random variable of $\mathcal{G}^t$ as $\mathbf{G}^t$. The prediction task on dynamic graphs aims to utilize past graph snapshots to make predictions, {\it i.e.}, $p(\mathbf{Y}^{t}|\mathbf{G}^{1:t})$, where $\mathbf{G}^{1:t}=\{\mathbf{G}^1,\mathbf{G}^2,\dots,\mathbf{G}^t\}$ denotes the graph trajectory, and the label $\mathbf{Y}^t$ represent the node properties or the links at time $t+1$. For brevity, we take node-level prediction tasks as an example in this paper. Following~\cite{zhang2022dynamic}, the distribution of graph trajectory can be factorized into ego-graph trajectories, such that $p(\mathbf{Y}^t \mid \mathbf{G}^{1:t})=\prod_v p(\mathbf{y}_v^t \mid \mathbf{G}_v^{1:t})$.

\paragraph{Distribution Shifts on Dynamic Graphs} 
The common optimization objective for prediction tasks on dynamic graphs is to learn an optimal predictor with empirical risk minimization (ERM), {\it i.e.} $\min_\theta \mathbb{E}_{(y^t,\mathcal{G}_v^{1:t}) \sim p_{tr}(\mathbf{y}^t,\mathbf{G}_v^{1:t})} \mathcal{L}(f_\theta(\mathcal{G}_v^{1:t}),y^{t})$,
where $f_\theta$ is a learnable dynamic graph neural networks. Under distribution shifts, however, the optimal predictor trained with ERM and the training distribution may not generalize well to the test distribution, since the risk minimization objectives under two distributions are different due to $p_{tr}(\mathbf{Y}^t,\mathbf{G}^{1:t})\neq p_{te}(\mathbf{Y}^t,\mathbf{G}^{1:t})$. The distribution shift on dynamic graphs is complex that may originate from temporal distribution shifts~\cite{gagnon2022woods,kim2021reversible,du2021adarnn,venkateswaran2021environment,lu2021diversify} as well as structural distribution shifts~\cite{wu2022discovering,wu2022handling,ding2021closer}. For example, trends or community structures can affect interaction patterns in co-author networks~\cite{jin2021community} and recommendation networks~\cite{wang2022causal}, {\it i.e.}, the distribution of ego-graph trajectories may vary through time and structures. 

Following out-of-distribution (OOD) generalization literature~\cite{ zhang2022dynamic,zhang2023outofdistribution,wu2022discovering, gagnon2022woods,arjovsky2019invariant,chang2020invariant, ahuja2020invariant}, we make the following assumptions of distribution shifts on dynamic graphs: 
\begin{assumption}
For a given task, there exists a predictor $f(\cdot)$, for samples ($\mathcal{G}^{1:t}_v$,$y^{t}$) from any distribution, there exists an invariant pattern $P^t_I(v)$ and a variant pattern $P^t_V(v)$  such that the following conditions are satisfied: 1) the invariant patterns are sufficient to predict the labels, $y^t_v=f(P^t_I(v))+\epsilon$, where $\epsilon$ is a random noise, 2) the observed data is composed of invariant and variant patterns, $P^t_I(v) = \mathcal{G}_v^{1:t} \backslash P^t_V(v)$, 3) the influence of the variant patterns on labels is shielded by the invariant patterns,  $\mathbf{y}_v^t \perp \mathbf{P}^t_V(v) \mid \mathbf{P}^t_I(v)$.
\label{assum:1}
\end{assumption} 

In the next section, inspired by~\cite{wu2022handling}, we give a motivation example to provide some high-level intuition before going to our formal method.

\section{Motivation Example}
Here we introduce a toy dynamic graph example to motivate learning invariant patterns in the spectral domain. We assume that the invariant and variant patterns lie in the 1-hop neighbors, {\it i.e.}, each node has an invariant subgraph and a variant subgraph. For simplicity, we focus on the number of neighbors, {\it i.e.}, each node $v$ has an invariant subgraph related degree $\mathbf{d}_{v,1} \in \mathbb{R}^{T\times 1}$ and a variant subgraph related degree $\mathbf{d}_{v,2} \in \mathbb{R}^{T\times 1}$. Only the former determines the node label, {\it i.e.}, $y_v = \mathbf{g}^\top \mathbf{d}_{v,1}$.
Note that invariant and variant subgraphs are not observed in the data.
We further assume a one-dimensional constant feature for each node, which is set as $1$ without loss of generality. 

For the model in the spatial-temporal domain, we adopt sum pooling as one-layer graph convolution, {\it i.e.}, the message passing for each node and time is $\mathbf{h}_v = \sum_{u \in \mathcal{N}_v} 1 = \mathbf{d}_{v,1}+ \mathbf{d}_{v,2}$. We further adopt a mask $\mathbf{m}\in \mathbb{R}^{T \times 1}$ to filter patterns in the temporal domain and make predictions by a linear classifier, {\it i.e.}, $\hat{y}_v = \mathbf{w}^\top(\mathbf{m} \odot \mathbf{h}_v)$, where $\mathbf{w} \in \mathbb{R}^{T\times 1}$ denotes the learnable parameters. Then, the empirical risk in the training dataset $D_{tr}$ is $R_{tr}(\mathbf{w})= \frac{1}{|D_{tr}|}\sum_{v\in D_{tr}}(\hat{y}_v - y_v)^2 $. We have the following proposition.

\begin{proposition}
For any mask $\mathbf{m}\in \mathbb{R}^{T \times 1}$, for the optimal classifier in the training data $\mathbf{w}^* = \argmin_{\mathbf{w}} R_{tr}(\mathbf{w})$ , if $||\mathbf{m}\odot\mathbf{w}^*||_2\neq 0$, there exist OOD nodes with unbounded error, {\it i.e.}, $\exists v$ s.t. $\lim_{||\mathbf{d}_{v,2}|| \to \infty}(\hat{y}_v - y_v)^2 = \infty$. 
\label{pro:1}
\end{proposition}
The proposition \ref{pro:1} shows that a classifier trained with masks and empirical risk minimization has unbounded risks in testing data under distribution shifts as the classifier uses variant patterns to make predictions. Next, we show that under mild conditions, an invariant linear classifier in the spectral domain can solve this problem. Denote $\mathbf{\Phi} \in \mathbb{C}^{T\times T}$ as the Fourier bases, where $\mathbf{\Phi}_{k,t} = \frac{1}{\sqrt{T}}e^{-j \frac{2\pi k t}{T} }$. 
Denote $\mathbf{z}_v = \mathbf{\Phi} \sum_{u\in\mathcal{N}_v} 1$ as the spectral representation after a linear message-passing.
The prediction is $\hat{y}_v = \mathbf{w}^{\text{H}}(\mathbf{m}\odot \mathbf{z}_{v})$, where $\mathbf{m}\in \mathbb{C}^{T\times 1}$ is the mask to filter the spectral patterns, $\mathbf{w} \in \mathbb{C}^{T \times 1}$ is a linear classifier, and $(\cdot)^{\text{H}}$ denotes Hermitian transpose. We have the following proposition.

\begin{proposition}
If $\Bigl(\overline{\mathbf{\Phi}\mathbf{d}_{v,1}} \odot \mathbf{\Phi}\mathbf{d}_{v,1} \Bigr) \odot \Bigl(\overline{\mathbf{\Phi}\mathbf{d}_{v,2}} \odot \mathbf{\Phi}\mathbf{d}_{v,2} \Bigr) =\mathbf{0}, \forall \mathbf{d}_{v,1}, \mathbf{d}_{v,2}$, then $ \exists 
\mathbf{m} \in \mathbb{C}^{T\times 1}$ such that the optimal spectral classifier in the training data has bounded error, {\it i.e.}, for $\mathbf{w}^* = \argmin_{\mathbf{w}} R_{tr}(\mathbf{w})$, $\exists \epsilon>0$, $\forall v, \lim_{||\mathbf{d_{v,2}}||\to \infty}(\hat{y}_v - y_v)^2 < \epsilon $.
\label{pro:2}
\end{proposition}

The proposition \ref{pro:2} shows that if the frequency bandwidths of invariant and variant patterns do not have any overlap, there exists a spectral mask such that a linear classifier trained with empirical risk minimization in the spectral domain will have bounded risk in any testing data distribution. This example motivates us to capture invariant and variant patterns in the spectral domain, which is not feasible in the spatial-temporal domain.

\section{Method}
\begin{figure}
\centering
\includegraphics[width=1\linewidth]{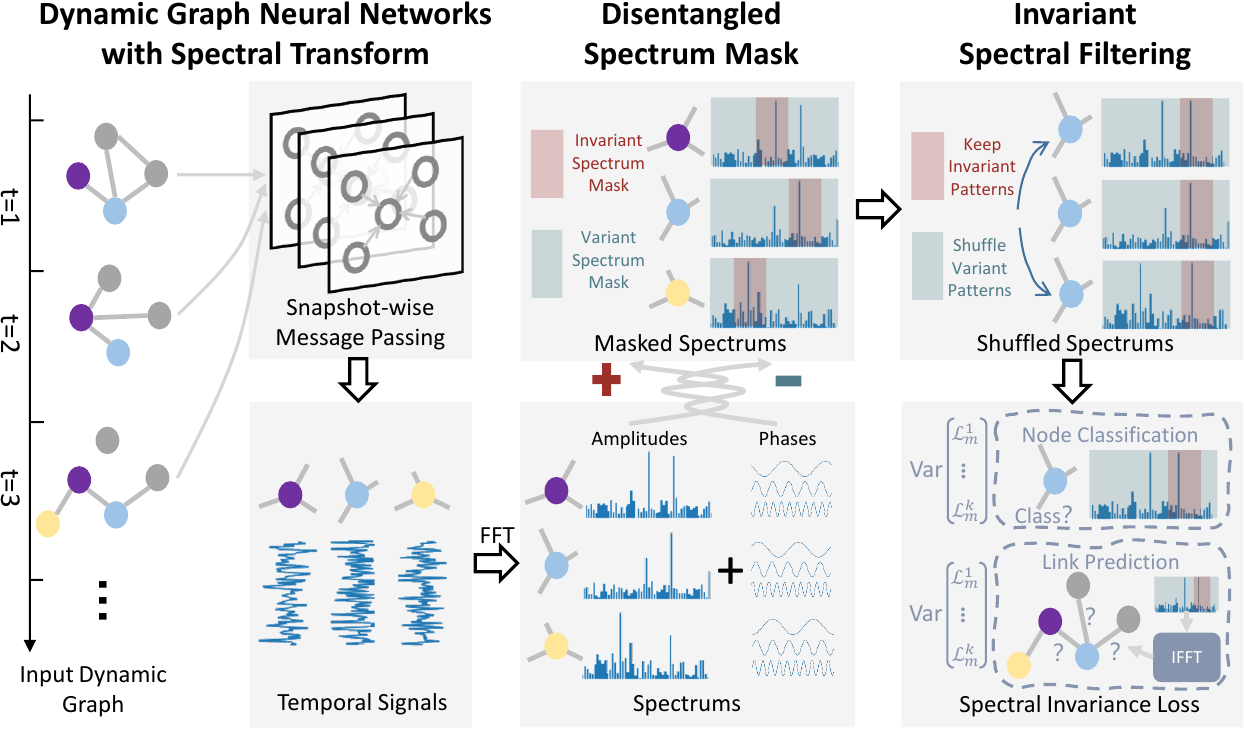}    
\caption{The framework of our proposed method \modelnosp. Given a dynamic graph evolving through time, the dynamic graph neural networks with spectral transform first obtain the ego-graph trajectory spectrums in the spectral domain. Then the disentangled spectrum mask leverages the amplitude and phase information of the ego-graph trajectory spectrums to obtain invariant and variant spectrum masks. Last, invariant spectral filtering discovers the invariant and variant patterns via the disentangled spectrum masks, and minimizes the variance of predictions with exposure to various variant patterns, to help the model exploit invariant patterns to make predictions under distribution shifts.}
\label{fig:framework}
\end{figure}

In this section, we introduce our method named Spectral Invariant Learning for Dynamic Graphs under Distribution Shifts (\modelnosp) to handle distribution shifts in dynamic graphs, including three modules, dynamic graph neural networks with Fourier transform, disentangled spectrum mask, and invariant spectral filtering.  
The framework of our method is shown in Figure~\ref{fig:framework}.

\subsection{Dynamic Graph Neural Network with Spectral Transform}
\paragraph{Dynamic Graph Trajectories Modeling}
Each node on the dynamic graph has its ego-graph trajectory evolving through time that may determine the node properties or the occurrence of future links. Following~\cite{you2022roland,sankar2020dysat,pareja2020evolvegcn}, we adopt a message-passing network for each graph snapshot to aggregate the neighborhood information at the current time, {\it i.e.}, 
\begin{equation}
\mathbf{m}^t_{u \rightarrow v} \leftarrow \mathrm{MSG}(\mathbf{h}^t_u, \mathbf{h}^t_v), 
\mathbf{h}^t_v \leftarrow\operatorname{AGG} (\{\mathbf{m}^t_{u \rightarrow v} \mid u \in \mathcal{N}^t(v)\}, \mathbf{h}^t_v\}),
\label{eq:gnn}
\end{equation}
where `MSG' and `AGG' denote message and aggregation functions, $\mathbf{m}^t_{u \rightarrow v}$ is the message from node $u$ to node $v$, $\mathbf{h}^t_u$ is the node embedding for node $u$ at time $t$, $\mathcal{N}^t(v) = \{u\mid (u,v) \in \mathcal{E}^t \}$ is node $v$'s neighborhood at time $t$. To model the high-order neighborhood information, we can stack multiple message-passing layers. In this way, the node embedding along time $\{\mathbf{h}^t_u\}^T_{t=1}$ summarizes the evolution of node $u$'s ego-graph trajectories. We denote $\mathbf{H} \in \mathbb{R}^{T \times N \times d}$ as the ego-graph trajectory signals for all nodes on the dynamic graph, where $T$ denotes the total time length, $N$ denotes the number of nodes and $d$ denotes the hidden dimensionality. 

\paragraph{Spectral Transform} As some patterns on dynamic graphs are unobservable in the time domain, while observable in the spectral domain, we transform the summarized ego-graph trajectory signals $\mathbf{H}$ into the spectral domain via Fourier transform for each node and hidden dimension, {\it i.e.},
\begin{equation}
\label{eq:fft}
\mathbf{\Phi}_{k,t} = \frac{1}{\sqrt{T}} e^{-j \frac{2\pi k t}{T}}, \mathbf{Z} = \mathbf{\Phi}\mathbf{H},
\end{equation}
where $\mathbf{\Phi} \in \mathbb{C}^{K \times T}$ denotes the Fourier bases, $K$ denotes the number of frequency components, and $\mathbf{Z}\in \mathbb{C}^{K \times N \times d}$ denote the node embeddings along frequency components in the spectral domain, and $\mathbf{Z}_{k,n,m} = \sum_{t=1}^{T} \mathbf{\Phi}_{k,t} \mathbf{H}_{t,n,m}$.  By choosing the Fourier bases, our spectral transform has the following advantages: 1) we can use fast Fourier transform (FFT)~\cite{cooley1965algorithm} to accelerate the computation. The computation complexity of Eq.~\eqref{eq:fft} can be reduced from $O(NdT^2)$ to $O(NdTlogT)$. 2) Each basis has clear semantics, {\it e.g.}, $\mathbf{Z}_k$ denotes the node embeddings at the $k$-th frequency component in the spectral domain. In this way, we can observe how the nodes on the dynamic graph evolve in different frequency bandwidths. 3) Fourier transform is able to capture global and periodic patterns~\cite{bracewell1986fourier, yi2023neural}, which are common in real-world dynamic graphs, {\it e.g.}, the interactions on e-commerce networks may result from seasonal sales or product service cycles. 

\subsection{Disentangled Spectrum Mask}
To capture invariant patterns in the spectral domain, we propose to explicitly learn spectrum masks to disentangle the invariant and variant patterns. The embeddings in the spectral domain contain the amplitude information as well as the phase information for each node
\begin{equation}
\text{Amp}(\mathbf{Z}) = \Bigl(\text{Imag}^2(\mathbf{Z}) + \text{Real}^2(\mathbf{Z}) \Bigr)^{\frac12} ,\phi(\mathbf{Z}) = \arctan \frac{\text{Imag}(\mathbf{Z})}{\text{Real}(\mathbf{Z})},
\label{eq:amp}
\end{equation}
where $\text{Real}(\cdot)$ and $\text{Imag}(\cdot)$ denote the real and imaginary part of the complex number, {\it i.e.}, $\mathbf{Z} = \text{Real}(\mathbf{Z}) + j \text{Imag}(\mathbf{Z})$, $j$ denotes the imaginary unit, $\text{Amp}(\mathbf{Z}) \in \mathbb{R}^{K\times N \times d}$ and $\phi(\mathbf{Z}) \in \mathbb{R}^{K\times N \times d}$ denote the amplitude and phase information respectively. For brevity, the tensor operators in Eq.~\eqref{eq:amp} are all element-wise, {\it e.g.}, $\text{(Imag}^2(\mathbf{Z}))_{i,j,k} = (\text{Imag}(\mathbf{Z}_{i,j,k}))^2$. Then, we obtain the spectrum masks by leveraging both the amplitude and phase information 
\begin{equation}
\label{eq:mask}
\mathbf{M} = \text{MLP}(\text{Real}(\mathbf{Z}) || \text{Imag}(\mathbf{Z})), \mathbf{M}_I = \text{sigmoid}(\mathbf{M}/\tau), \mathbf{M}_V = \text{sigmoid}(-\mathbf{M}/\tau),
\end{equation}
where $\text{MLP}$ denotes multi-layer perceptrons, $\tau$ is the temporature, $\textbf{M}_I \in [0,1]^{K}$ and $\textbf{M}_V \in [0,1]^{K}$ denote the spectrum mask for invariant and variant patterns, and $||$ represents the concatenation of the embeddings. In this way, the invariant and variant masks have a negative relationship, and each node can have its own spectrum mask. As the phase information includes high-level semantics in the original signals~\cite{xu2021fourier,hansen2007structural,oppenheim1981importance,piotrowski1982demonstration, oppenheim1979phase}, we keep the phase information unchanged to reduce harm in the fine-grained semantic information for the graph trajectories, and filter the spectrums by the learned disentangled masks in terms of amplitudes,
\begin{equation}
\small
\label{eq:filter}
\mathbf{Z}_I = \Bigl(\mathbf{M}_I \odot \text{Amp}(\mathbf{Z})\Bigr) \odot (\cos \phi(\mathbf{Z}) + j \sin \phi(\mathbf{Z})), \mathbf{Z}_V = \Bigl(\mathbf{M}_V \odot \text{Amp}(\mathbf{Z})\Bigr)(\cos \phi(\mathbf{Z}) + j \sin \phi(\mathbf{Z})),
\end{equation}
where $\mathbf{Z}_I$ and $\mathbf{Z}_V$ denote the summarized invariant and variant patterns in the spectral domain. 
For node classification tasks, we can directly adopt the spectrums for the classifier to predict classes. For link prediction tasks, we can utilize inverse fast Fourier transform (IFFT) to transform the embeddings into the temporal domain for future link prediction
\begin{equation}
\label{eq:invfft}
\mathbf{H}^\prime_I = \mathbf{\Phi}^{\text{H}}\mathbf{Z}_I, \mathbf{H}^\prime_V = \mathbf{\Phi}^{\text{H}}\mathbf{Z}_V,
\end{equation}
where $(\cdot)^{\text{H}}$ is Hermitian transpose, $\mathbf{H}^\prime_I$ and $\mathbf{H}^\prime_V$ denote the filtered invariant and variant patterns that are transformed back into the temporal domain respectively. 

\subsection{Invariant Spectral Filtering} 
Under distribution shifts, the variant patterns on dynamic graphs have varying relationships with labels, while the invariant patterns have sufficient predictive abilities with regard to labels. 
We propose invariant spectral filtering to capture the invariant and variant patterns in the spectral domain, and help the model focus on invariant patterns to make predictions, thus handling distribution shifts. We take node classification tasks for an example as follows. 

Let $\mathbf{Z}_I \in \mathbb{C}^{K\times N \times d}$ and $\mathbf{Z}_V \in \mathbb{C}^{K\times N \times d}$ be the filtered invariant and variant spectrums in the spectral domain. 
Then we can utilize the invariant and variant node spectrums to calculate the task loss
\begin{equation}
\label{eq:task_loss}
\mathcal{L}_I = l(f_I(\mathbf{Z}_I), \mathbf{Y}), \mathcal{L}_V = l(f_V(\mathbf{Z}_V), \mathbf{Y}),
\end{equation}
where $f_I(\cdot)$ and $f_V(\cdot)$ are the classifiers for invariant and variant patterns respectively, $\mathbf{Y}$ is the labels, and $l$ is the loss function. The task loss is utilized to capture the patterns with the predictive abilities of labels. Recall in Assumption~\ref{assum:1}, the influence of variant patterns on labels is shielded given invariant patterns as the invariant patterns have sufficient predictive abilities w.r.t labels, and thus the model's predictions should not change when being exposed to different variant patterns and the original invariant patterns. Inspired by~\cite{wu2022discovering,zhang2022dynamic}, we calculate the invariance loss by  
\begin{equation}
\label{eq:invariance_loss}
\mathcal{L}_{INV} = \text{Var}(\{\mathcal{L}_m \mid \tilde{\mathbf{z}} : \tilde{\mathbf{z}} \in \mathcal{S} \}),
\end{equation}
where $\mathcal{L}_m \mid \tilde{\mathbf{z}}$ denotes the mixed loss to measure the model's prediction ability with exposure to the specific variant pattern $\tilde{\mathbf{z}} \in \mathbb{C}^{K\times d}$ that is sampled from a set of variant patterns $\mathcal{S}$. We adopt all the node embeddings in $\mathbf{Z}_V$ to construct the set of variant patterns $\mathcal{S}$. Inspired by~\cite{cadene2019rubi, wu2022discovering}, we calculate the mixed loss as 
\begin{equation}
\mathcal{L}_m \mid \tilde{\mathbf{z}} = l(f_I(\mathbf{Z}_I) \odot \sigma(f_V(\tilde{\mathbf{z}})), \mathbf{Y}),
\end{equation}
where $\sigma$ denotes the sigmoid function. Then, the final training objective is 
\begin{equation}
\label{eq:objective}
\min_{\theta,f_I} \mathcal{L}_I + \lambda\mathcal{L}_{INV} + \min_{f_V} \mathcal{L}_V,
\end{equation}
where $\theta$ is the parameters that encompass all the model parameters except the classifiers, $\lambda$ is a hyperparameter to balance the trade-off between the model's predictive ability and invariance properties. A larger $\lambda$ encourages the model to capture patterns with better invariance under distribution shifts, with the potential risk of lower predictive ability during training, as the shortcuts brought by the variant patterns might be discarded in the training process. After training, we only adopt invariant patterns to make predictions in the inference stage. The overall algorithm for training on node classification datasets is summarized in Algo.~\ref{algo:pipeline}.

\begin{algorithm}
\caption{Training pipeline for \model on node classification datasets} 
\label{algo:pipeline}
\begin{algorithmic}[1]
\REQUIRE 
Training epochs $L$, sample number $S$, hyperparameter $\lambda$
\FOR{$l = 1, \dots, L$}
    \STATE Obtain the node embeddings $\mathbf{H}$ with snapshot-wise message passing as Eq.~\eqref{eq:gnn}
    \STATE Transform the node embeddings into the spectral domain with FFT as Eq.~\eqref{eq:fft}
    \STATE Calculate the disentangled spectrum masks as Eq.~\eqref{eq:mask}
    \STATE Filter spectrums into invariant and variant patterns as Eq.~\eqref{eq:filter}
    \STATE Calculate the task loss as Eq.~\eqref{eq:task_loss}
    \STATE Sample $S$ variant patterns from collections of $\mathbf{Z}_V$ and calculate the invariance loss as Eq.~\eqref{eq:invariance_loss}
    \STATE Update the model according to Eq.~\eqref{eq:objective}
\ENDFOR
\end{algorithmic}
\end{algorithm}

\section{Experiments}
In this section, we conduct extensive experiments to verify that our proposed method can handle distribution shifts on dynamic graphs by discovering and utilizing invariant patterns in the spectral domain. More details of the settings and other results can be found in Appendix. 

\textbf{Baselines.} We adopt several representative dynamic GNNs and Out-of-Distribution(OOD) generalization methods as our baselines:

\begin{itemize}[leftmargin=0.5cm]    
    \item Dynamic GNNs: 
    \textbf{GCRN}~\cite{seo2018structured} is a representative dynamic GNN that first adopts a GCN\cite{kipf2016variational} to obtain node embeddings and then a GRU~\cite{Cho2014LearningPR} to model the network evolution.
    \textbf{EGCN}~\cite{pareja2020evolvegcn} adopts an LSTM~\cite{hochreiter1997long} or GRU~\cite{Cho2014LearningPR} to flexibly evolve the GCN~\cite{kipf2016variational} parameters through time.
    \textbf{DySAT}~\cite{sankar2020dysat} aggregates neighborhood information at each graph snapshot using structural attention and models network dynamics with temporal self-attention.
    
    \item OOD generalization methods: \textbf{IRM}~\cite{arjovsky2019invariant} aims at learning an invariant predictor which minimizes the empirical risks for all training domains.
    \textbf{GroupDRO}~\cite{sagawa2019distributionally} puts more weight on training domains with larger errors to minimize the worst-group risks across training domains.
    \textbf{V-REx}~\cite{krueger2021out} reduces the differences in the risks across training domains to reduce the model’s sensitivity to distributional shifts. As these methods are not specifically designed for dynamic graphs, we adopt the best-performed dynamic GNNs as their backbones on each dataset.

    \item OOD generalization methods for dynamic graphs: 
    \textbf{DIDA}~\cite{zhang2022dynamic} utilizes disentangled attention to capture invariant and variant patterns in the spatial-temporal domain, and conducts spatial-temporal intervention mechanism to let the model focus on invariant patterns to make predictions.   
    
\end{itemize}

\begin{table}[]
\centering
\caption{Results of different methods on real-world link prediction and node classification datasets. The best results are in bold and the second-best results are underlined.  The year in the Aminer dataset denotes the test split, {\it e.g.}, `Aminer15' denotes the average test accuracy in 2015.}
\adjustbox{max width=\textwidth}{
\begin{tabular}{cccccc}
\toprule
\textbf{Task}    & \multicolumn{2}{c}{\textbf{Link Prediction (AUC\%)}} & \multicolumn{3}{c}{\textbf{Node Classification (ACC\%)}}        \\
\textbf{Dataset} & \textbf{Collab}           & \textbf{Yelp}            & \textbf{Aminer15}   & \textbf{Aminer16}   & \textbf{Aminer17}   \\ \midrule
GCRN             & \ms{69.72}{0.45}          & \ms{54.68}{7.59}         & \ms{47.96}{1.12}    & \ms{51.33}{0.62}    & \ms{42.93}{0.71}    \\
EGCN             & \ms{76.15}{0.91}          & \ms{53.82}{2.06}         & \ms{44.14}{1.12}    & \ms{46.28}{1.84}    & \ms{37.71}{1.84}    \\
DySAT            & \ms{76.59}{0.20}          & \ms{66.09}{1.42}         & \ms{48.41}{0.81}    & \ms{49.76}{0.96}    & \ms{42.39}{0.62}    \\
IRM              & \ms{75.42}{0.87}          & \ms{56.02}{16.08}        & \ms{48.44}{0.13}    & \ms{50.18}{0.73}    & \ms{42.40}{0.27}    \\
VREx             & \ms{76.24}{0.77}          & \ms{66.41}{1.87}         & \ms{48.70}{0.73}    & \ms{49.24}{0.27}    & \ms{42.59}{0.37}    \\
GroupDRO         & \ms{76.33}{0.29}          & \ms{66.97}{0.61}         & \ms{48.73}{0.61}    & \ms{49.74}{0.26}    & \ms{42.80}{0.36}    \\
DIDA             & \mstwo{81.87}{0.40}       & \mstwo{75.92}{0.90}      & \mstwo{50.34}{0.81} & \mstwo{51.43}{0.27} & \mstwo{44.69}{0.06} \\ \midrule
\model           & \msone{84.09}{0.16}       & \msone{78.65}{2.22}      & \msone{52.35}{1.04} & \msone{54.11}{0.62} & \msone{45.54}{1.19} \\ \bottomrule
\end{tabular}
}
\label{tab:real}
\end{table}

\begin{table}[]
\centering
\caption{Results of different methods on synthetic link prediction and node classification datasets. The best results are in bold and the second-best results are underlined. A larger `shift' denotes a higher distribution shift level.}
\adjustbox{max width=\textwidth}{
\begin{tabular}{ccccccc}
\toprule
\textbf{Dataset} & \multicolumn{3}{c}{\textbf{Link-Synthetic (AUC\%)}}             & \multicolumn{3}{c}{\textbf{Node-Synthetic (ACC\%)}}             \\
\textbf{Shift}   & \textbf{0.4}                 & \textbf{0.6}                 & \textbf{0.8}                 & \textbf{0.4}                 & \textbf{0.6}                 & \textbf{0.8}                 \\ \midrule
GCRN             & \ms{72.57}{0.72}    & \ms{72.29}{0.47}    & \ms{67.26}{0.22}    & \ms{27.19}{2.18}    & \ms{25.95}{0.80}    & \ms{29.26}{0.69}    \\
EGCN             & \ms{69.00}{0.53}    & \ms{62.70}{1.14}    & \ms{60.13}{0.89}    & \ms{24.01}{2.29}    & \ms{22.75}{0.96}    & \ms{24.98}{1.32}    \\
DySAT            & \ms{70.24}{1.26}    & \ms{64.01}{0.19}    & \ms{62.19}{0.39}    & \ms{40.95}{2.89}    & \ms{37.94}{1.01}    & \ms{30.90}{1.97}    \\
IRM              & \ms{69.40}{0.09}    & \ms{63.97}{0.37}    & \ms{62.66}{0.33}    & \ms{33.23}{4.70}    & \ms{30.29}{1.71}    & \ms{29.43}{1.38}    \\
VREx             & \ms{70.44}{1.08}    & \ms{63.99}{0.21}    & \ms{62.21}{0.40}    & \ms{41.78}{1.30}    & \ms{38.11}{2.81}    & \ms{29.56}{0.44}    \\
GroupDRO         & \ms{70.30}{1.23}    & \ms{64.05}{0.21}    & \ms{62.13}{0.35}    & \ms{41.35}{2.19}    & \ms{35.74}{3.93}    & \mstwo{31.03}{1.24} \\
DIDA             & \mstwo{85.20}{0.84} & \mstwo{82.89}{0.23} & \mstwo{72.59}{3.31} & \mstwo{43.33}{7.74} & \mstwo{39.48}{7.93} & \ms{28.14}{3.07}    \\ \midrule
\model           & \msone{85.95}{0.18} & \msone{84.69}{1.18} & \msone{78.01}{0.71} & \msone{43.62}{2.74} & \msone{39.78}{3.56} & \msone{38.64}{2.76} \\ \bottomrule
\end{tabular}
}
\label{tab:synthetic}
\end{table}

\subsection{Real-world Datasets}
\paragraph{Settings} We use 3 real-world dynamic graph datasets, including Collab~\cite{Tang:12KDDCross, zhang2022dynamic}, Yelp~\cite{sankar2020dysat, zhang2022dynamic} and Aminer~\cite{tang2008arnetminer,sinha2015overview}. 
Following~\cite{zhang2022dynamic}, we adopt the challenging inductive future link prediction task on Collab and Yelp, where the model should exploit historical graphs to predict the occurrence of links in the next time step. To measure the model's performance under distribution shifts, the model is tested on another dynamic graph with different fields, which is unseen during training. 
For node classification, we adopt Aminer, a citation network, where nodes represent papers, and edges from $u$ to $v$ with timestamp $t$ denote the paper $u$ published at year $t$ cites the paper $v$. The task is to predict the venues of the papers. We train on papers published between 2001 - 2011, validate on those published in 2012 - 2014, and test on those published since 2015. On this dataset, the model is tested to exploit the invariant patterns and make stable predictions under distribution shifts, where the patterns on the dynamic graph may vary in different years. 

\paragraph{Results} Based on the results in Table~\ref{tab:real}, we have the following observations: 1) {\it Under distribution shifts, the general OOD generalization baselines have limited improvements over the dynamic GNNs}, {\it e.g.}, GroupDRO improves over DySAT with 0.9\% in Yelp and 0.3\% in Aminer15 respectively. A plausible reason is that they are not specially designed to handle distribution shifts on dynamic graphs, and may not consider the graph structural and temporal dynamics to capture invariant patterns. Another reason might be that they strongly rely on high-quality environment labels to capture invariant patterns, which are almost unavailable on real-world dynamic graphs. 2) {\it Our method can better handle distribution shifts than the baselines.} The datasets have strong distribution shifts, {\it e.g.}, COVID-19 happens midway and has considerable influence on the consumer behavior on Yelp, and the citation patterns may shift with the outbreak of deep neural networks on Aminer. Nevertheless, our method \model has significant improvements over the state-of-the-art OOD generalization baseline for dynamic graphs DIDA on all datasets, {\it e.g.}, 2\% on average for most datasets, which verifies that our method can better capture the invariant and variant patterns in the spectral domain, and thus handling distribution shifts on dynamic graphs.  

\subsection{Synthetic Datasets}
\paragraph{Settings} To evaluate the model's generalization ability under distribution shifts, we conduct experiments on synthetic link prediction and node classification datasets, which are constructed by introducing manually-designed distribution shifts. 
For link prediction datasets, we follow~\cite{zhang2022dynamic} to generate additional varying features for each node and timestamps on the original dataset Collab, where these additional features are constructed with spurious correlations w.r.t the labels, {\it i.e.}, the links in the next timestamps. The spurious correlation degree is determined by a shift level parameter. On this dataset, to have better generalization ability, the model should not rely on variant patterns that exploit the additional features with spurious correlations.
For node classification, we briefly introduce the construction of the synthetic dataset as follows. We generate the dynamic graph with stochastic block model~\cite{HOLLAND1983109}, where the link 
probability between nodes at each graph snapshot is determined by two frequency factors. The correlation of one of the factors with class labels is always 1, while the other factor has a variant relationship with labels, where the relationship is also controlled by a shift level parameter. The model should discover and focus on the invariant frequency factors whose relationship with labels is invariant under distribution shifts.  
For both datasets, we set the shift level parameters as 0.4, 0.6, 0.8 for training and validation splits, and 0 for test splits. 

\paragraph{Results} Based on the results in Table~\ref{tab:synthetic}, we have the following observations: 1) {\it Our method can better handle distribution shifts than the baselines, especially under stronger distribution shifts.} \model consistently outperforms DyGNN and general OOD generalization baselines by a significantly large margin, which can credit to our special design to handle distribution shifts on dynamic graphs in the spectral domain. Our method also has a significant improvement over the best-performed baseline under the strongest distribution shift, {\it e.g.}, with absolute improvements of 5\% in Link-Synthetic(0.8) and 7\% in Node-Synthetic(0.8) respectively. 2) {\it Our method can exploit invariant patterns to consistently alleviate the harmful effects of variant patterns under different distribution shift levels.} As the distribution shift level increases, almost all methods decline in performance since the relationship between variant patterns and labels goes stronger, so that the variant patterns are much easier to be exploited by the model, misleading the training process. However, the performance drop of \model is significantly lower than baselines, which demonstrates that our method can alleviate the harmful effects of variant patterns under distribution shifts by exploiting invariant patterns in the spectral domain.

\subsection{Ablation Studies}

\begin{figure}
\centering
\includegraphics[width=0.95\textwidth]{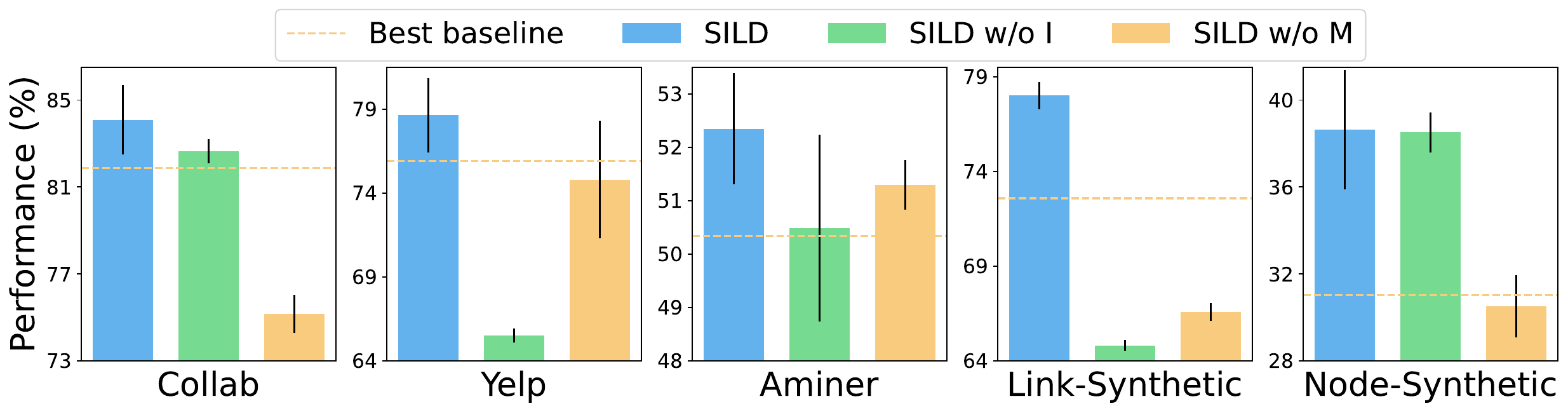}    
\caption{Results of ablation studies, where `w/o I' removes invariant spectral filtering in \model, `w/o M' removes disentangled spectrum masks, and `Best baseline' denotes the best-performed baseline on each dataset. The error bars report the standard deviations. (Best viewed in color)}
\label{fig:ablation}
\vspace{-5pt}
\end{figure}

We conduct ablation studies to verify the effectiveness of the proposed disentangled spectrum mask and invariant spectral filtering in \modelnosp. The ablated version `SILD w/o I' removes invariant spectral filtering in \model by setting $\lambda = 0$, and `SILD w/o M' is trained without the disentangled spectrum masks. From Figure~\ref{fig:ablation}, we have the following observations. First, our proposed \model outperforms all the variants as well as the best-performed baseline on all datasets, demonstrating the effectiveness of each component of our proposed method. Second, `SILD w/o I' and `SILD w/o M' drop drastically in performance on all datasets compared to the full version, which verifies that our proposed disentangled spectrum mask and spectral invariant learning can help the model to focus on invariant patterns to make predictions and significantly improve the performance under distribution shifts.  

\section{Related Works}
\paragraph{Dynamic Graph Neural Networks} Dynamic graphs ubiquitously exist in real-world applications~\cite{cai2021structural,deng2020dynamic,you2019hierarchical,wang2021tedic,li2019fates,wu2020temp, zhang2023dynamic,zhang2023ood, zhang2023LLM4DyG} such as event forecasting, recommendation, etc. In comparison with static graphs~\cite{zhang2020deep,zhang2023unsupervised,guan2021autogl,qin2022bench,qin2021graph,zhang2023large,wang2017community,wang2019heterogeneous}, dynamic graphs contain rich temporal information. Considerable research attention has been devoted to dynamic graph neural networks (DyGNNs)~\cite{skarding2021foundations,zhu2022learnable,chen2023easydgl} to model the complex graph dynamics that include structures and features evolving through time. Some works adopt GNN to aggregate neighborhood information for each graph snapshot, and then utilize a sequence module to model the temporal information~\cite{yang2021discrete,sun2021hyperbolic,hajiramezanali2019variational,seo2018structured,sankar2020dysat}.  Some others utilize time-encoding techniques to encode the temporal links into time-aware embeddings and adopt a GNN or memory module~\cite{wang2021inductive,cong2021dynamic,xu2020inductive,rossi2020temporal} to process structural information. Some other related works leverage spectral graph neural networks~\cite{cao2020spectral}, global graph framelet convolution~\cite{zhou2022well}, and graph wavelets~\cite{bastos2022learnable} to obtain better dynamic graph representations. 
However, distribution shifts remain largely unexplored in dynamic graph neural networks literature. 
The sole prior work DIDA~\cite{zhang2022dynamic} handles spatial-temporal distribution shifts on dynamic graphs in the temporal domain. To the best of our knowledge, this is the first study of handling distribution shifts on dynamic graphs in the spectral domain. 

\paragraph{Out-of-Distribution Generalization}
A significant proportion of existing machine learning methodologies operate on the assumption that training and testing data are independent and identically distributed (i.i.d.). However, this assumption may not always hold true, especially in the context of complex real-world scenarios~\cite{shen2021towards}, and the uncontrollable distribution shifts between training and testing data distribution may lead to a significant decline in the model performance. Out-of-Distribution (OOD) generalization problem has recently drawn great attention in various areas~\cite{zhou2021domain,shen2021towards,yao2022improving}. Some works handle structural distribution shifts on static graphs~\cite{li2022out,wu2022discovering,wu2022handling,chen2022invariance,zhu2021shift,qin2022graph,li2022ood,zhang2022learning,zhang2021revisiting,fan2021generalizing,li2022gil} and temporal distribution shifts on time-series data~\cite{gagnon2022woods,du2021adarnn,kim2021reversible,venkateswaran2021environment,lu2021diversify,yao2022wildtime}. However, how to handle distribution shifts on dynamic graphs in the spectral domain remains unexplored. 

\paragraph{Spectral Methods in Neural Networks} The applications of spectral methods in neural networks have been broadly explored in many areas, including static graph data~\cite{defferrard2016convolutional, xu2019graph, kenlay2021interpretable, wang2022powerful, bo2023specformer}, time-series data~\cite{cao2020spectral, salomone2020spectral, lange2021fourier, zhou2022fedformer,zhang2022self}, etc., for their advantages of modeling global patterns, powerful expressiveness and interpretability~\cite{bo2023survey, yi2023neural}. Some work~\cite{jiang2021focal} proposes to reconstruct the image in the spectral domain to obtain robust image representations. Some work~\cite{xu2021fourier} proposes to augment the image data by perturbing the amplitude information in the spectral domain. Some work~\cite{gupta2021multiwavelet} proposes a multiwavelet-based method for compressing operator kernels. However, these methods are not applicable to dynamic graphs, not to mention the more complex scenarios under distribution shifts.  

\section{Conclusion}
In this paper, we propose a novel model named Spectral Invariant Learning for Dynamic Graphs under Distribution Shifts (\modelnosp), which can handle distribution shifts on dynamic graphs in the spectral domain. We design a DyGNN with Fourier transform to obtain the ego-graph trajectory spectrums. Then we propose a disentangled spectrum mask and invariant spectral filtering to discover the invariant and variant patterns in the spectral domain, and help the model rely on invariant spectral patterns to make predictions. Extensive experimental results on several synthetic and real-world datasets, including both node classification and link prediction tasks, demonstrate the superior performance of our method compared to state-of-the-art baselines under distribution shifts. \red{One limitation is that in this paper we mainly focus on dynamic graphs in scenarios of discrete snapshots, and we leave extending our methods to continuous dynamic graphs for further explorations.}

\section*{Acknowledgements}

This work was supported by the National Key Research and Development Program of China No. 2020AAA0106300, National Natural Science Foundation of China (No. 62222209, 62250008, 62102222, 62206149), Beijing National Research Center for Information Science and Technology under Grant No. BNR2023RC01003, BNR2023TD03006, China National Postdoctoral Program for Innovative Talents No. BX20220185, China Postdoctoral Science Foundation No. 2022M711813, and Beijing Key Lab of Networked Multimedia. All opinions, findings, conclusions, and recommendations in this paper are those of the authors and do not necessarily reflect the views of the funding agencies.

\medskip
{
\small
\bibliographystyle{unsrt}
\bibliography{main}
}

\appendix

\section{Notations}
\begin{table}[ht]
\small
\caption{The summary of the notations and their descriptions }
\begin{tabular}{c|l}
\toprule
\textbf{Notations}                                                  & \textbf{Descriptions}                                                                   \\ \midrule
$\mathcal{G}=(\mathcal{V},\mathcal{E})$                    & A graph with the node set and edge set                                         \\
$\mathcal{G}^t=(\mathcal{V}^t,\mathcal{E}^t)$              & Graph slice at time $t$                                                        \\
$\mathbf{X}^t,\mathbf{A}^t$                                    & Features and adjacency matrix of a graph at time $t$                                       \\
$\mathcal{G}^{1:t},Y^t, \mathbf{G}^{1:t},\mathbf{Y}^t$     & Graph trajectory, label and their corresponding random variable                \\
$\mathcal{G}_v^{1:t},y_v^t, \mathbf{G}_v^{1:t},\mathbf{y}_v^t$ & Ego-graph trajectory, the node's label and their corresponding random variable \\
$p(\cdot)$ & Probability distribution\\
$P,\mathbf{P}$                                             & Pattern and its corresponding random variable                                  \\
% $\epsilon$                                             & Random noise                                  \\
$\mathbf{d}_{v,1}, \mathbf{d}_{v,2}$                        & The degrees of node $v$ varying by time                                                    \\
$\mathbf{g},\mathbf{w}$ & The parameters of linear classifiers \\
$\mathbf{m},\mathbf{M}$ & The mask to filter node representations \\
$\hat{y}_v$ & The prediction for the node $v$ \\
$R_{tr}(\mathbf{w})$ & The risks of the classifier $\mathbf{w}$ in training data  \\
$\mathbf{\Phi}$ & The Fourier bases \\
$\overline{x}$ & The conjugate of $x$  \\
$\text{MSG}(\cdot),\text{AGG}(\cdot)$                            & Message and Aggregation functions                       \\
$\mathbf{h}_u^t$                                           & Hidden embeddings for node $u$ at time $t$                                     \\
$\mathbf{H},\mathbf{Z}$                                           & Node representations in the temporal domain and spectral domain                                   \\
$d$                                                         & The dimensionality of node representations                                      \\
$\text{Amp}(\mathbf{Z}), \phi(\mathbf{Z})$ & The amplitudes and phases of the representations $\mathbf{Z}$  \\
$\mathbf{x}^H$ & The Hermitian transpose of $\mathbf{x}$ \\
$T,K$ &  The number of time stamps and the number of frequency components \\
$f(\cdot)$                                        & Predictors     \\             
$\ell$                                                     & Loss function                                                    \\
$\mathcal{L},\mathcal{L}_m,\mathcal{L}_{INV}$                 & Task loss, mixed loss and invariance loss                                      \\ \bottomrule
\end{tabular}
\end{table}

\section{Theoretical Analyses}
% \newpage
% \appendix
% \section{Notations}

\subsection{Proof of Proposition~\ref{pro:1}} 
\begin{proof}
For any mask $\mathbf{m}\in \mathbb{R}^{T\times 1}$, the predictions of the model is
\begin{equation}
\hat{y}_v = \mathbf{w}^\top (\mathbf{m}\odot(\mathbf{d}_{v,1} + \mathbf{d}_{v,2})).
\end{equation}

We assume that $||\mathbf{m}\odot \mathbf{w}||\neq 0$, otherwise the classifier is a trivial solution and always predicts $\hat{y}_v = 0$ for any node $v$. The empirical risk in training data is
\begin{align}
R_{tr}(\mathbf{w})&= \frac{1}{|D_{tr}|}\sum_{v \in D_{tr}} (\hat{y}_v - y_v)^2.
\end{align}

By setting $\frac{\partial R_{tr}(\mathbf{w})}{\partial 
 \mathbf{w}} =0$, we have the optimal classifier learned from the training data
\begin{align}
\mathbf{w}^* &= \frac{ \mathbf{m} \odot \sum_{v\in D_{tr}} (\mathbf{d}_{v,1} + \mathbf{d}_{v,2})\mathbf{g}^\top\mathbf{d}_{v,1}}
{\sum_{v\in D_{tr}} (\mathbf{m} \odot(\mathbf{d}_{v,1} + \mathbf{d}_{v,2}))\top (\mathbf{m} \odot(\mathbf{d}_{v,1} + \mathbf{d}_{v,2}))}.
\end{align}

Then for a node $v$ that has variant patterns  $\mathbf{d}_{v,2} = \alpha \mathbf{m}\odot \mathbf{w}^*$ and $\alpha \in \mathbb{R}$, the loss of the model's prediction is  
\begin{equation}
\begin{aligned}
l_v = (\hat{y}_v - y_v)^2 &= \Bigl({\mathbf{w}^*}^\top (\mathbf{m}\odot(\mathbf{d}_{v,1}+\mathbf{d}_{v,2} )) - \mathbf{g}^\top\mathbf{d}_{v,1} \Bigr)^2
\\&= \Bigl( \alpha|| \mathbf{m}\odot \mathbf{w}^*||^2 + ({\mathbf{w}^*}^\top \mathbf{m}\odot\mathbf{d}_{v,1} - \mathbf{g}^\top\mathbf{d}_{v,1}) \Bigr)^2
\\&= \Bigl( ||\mathbf{d}_{v,2}|| ||\mathbf{m}\odot \mathbf{w}^*|| + ({\mathbf{w}^*}^\top \mathbf{m}\odot\mathbf{d}_{v,1} - \mathbf{g}^\top\mathbf{d}_{v,1}) \Bigr)^2.
\end{aligned}
\end{equation}

Then $\forall \epsilon > 0$, when $||\mathbf{d}_{v,2}||> \frac{\sqrt{\epsilon} - ({\mathbf{w}^*}^\top \mathbf{m}\odot\mathbf{d}_{v,1} - \mathbf{g}^\top\mathbf{d}_{v,1})}{ ||\mathbf{m}\odot \mathbf{w}^* ||}$, i.e., $\alpha > \frac{\sqrt{\epsilon} - ({\mathbf{w}^*}^\top \mathbf{m}\odot\mathbf{d}_{v,1} - \mathbf{g}^\top\mathbf{d}_{v,1})}{ ||\mathbf{m}\odot \mathbf{w}^* ||^2}$, $(\hat{y}_v - y_v)^2 > \epsilon$, indicating that $\lim_{||\mathbf{d}_{v,2}|| \to \infty}(\hat{y}_v - y_v)^2 = \infty$.
Thus we conclude the proof.
\end{proof}

\subsection{Proof of Proposition~\ref{pro:2}}
\begin{proof}
Let the mask in the spectral domain $\mathbf{m} \in \mathbb{C}^{K\times 1}$ be
\begin{equation}
\mathbf{m}_i =
\begin{cases}
0 & \text{ if }  \exists v, \Bigl(\overline{\mathbf{\Phi}\mathbf{d}_{v,2}} \odot  \mathbf{\Phi}\mathbf{d}_{v,2} \Bigr)_i \neq 0 \\
1 & \text{otherwise}
\end{cases}.    
\end{equation}

Since the frequency bandwidths of invariant and variants patterns do not have overlap, i.e., $\Bigl(\overline{\mathbf{\Phi}\mathbf{d}_{v,1}} \odot \mathbf{\Phi}\mathbf{d}_{v,1} \Bigr) \odot \Bigl(\overline{\mathbf{\Phi}\mathbf{d}_{v,2}} \odot \mathbf{\Phi}\mathbf{d}_{v,2} \Bigr) =\mathbf{0}, \forall \mathbf{d}_{v,1}, \mathbf{d}_{v,2}$, we have $\mathbf{m}_i\odot \mathbf{z}_{v,1} = \mathbf{z}_{v,1}$ and $\mathbf{m}_i\odot \mathbf{z}_{v,2}$ for any node $v$. 
Let $\mathbf{w}_1 = \mathbf{\Phi}\mathbf{g}$, then the prediction for any node $v$ is 
\begin{equation}
\begin{aligned}
\hat{y}_v &= (\mathbf{\Phi}\mathbf{g})^{\text{H}} (\mathbf{m}\odot(\mathbf{z}_{v,1}+\mathbf{z}_{v,2}))
\\&= \mathbf{g}^\top \mathbf{\Phi}^{\text{H}} (\mathbf{z}_{v,1})
\\&= \mathbf{g}^\top \mathbf{\Phi}^{\text{H}} (\mathbf{\Phi} \mathbf{d}_{v,1})
\\&= \mathbf{g}^\top \mathbf{d}_{v,1}.
\end{aligned}
\end{equation}
For any node $v$, we have $(\hat{y}_v - y_v)^2 =0 $, so that $\mathbf{w}_1 = \argmin_{\mathbf{w}} R_{tr}(\mathbf{w})$, and $\forall v, \lim_{||\mathbf{d_{v,2}}||\to \infty}(\hat{y}_v - y_v)^2 < 1 $. Thus we conclude the proof.
\end{proof}

\section{Additional Experiments and Analyses}

\subsection{Hyperparameter Sensitivity}
\begin{figure}
\centering
\includegraphics[width=0.9\textwidth]{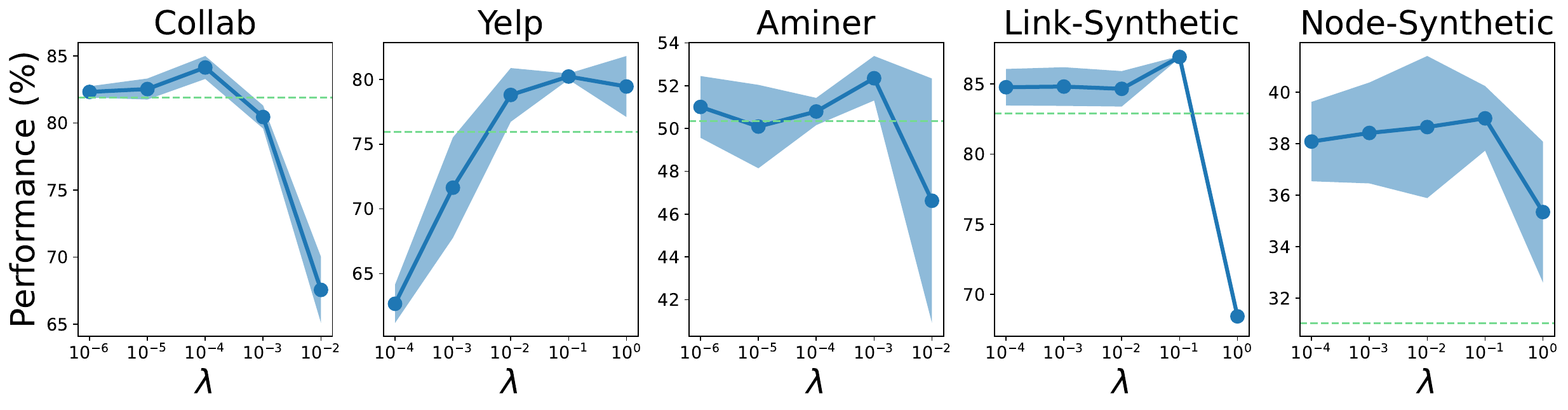}    
\caption{Sensitivity of hyperparameter $\lambda$. The area shows the average AUC and standard deviations in the test stage. The dashed line represents the average AUC of the best performed baseline.}
\label{fig:sens}
\end{figure}

We analyze the sensitivity of hyperparameter $\lambda$ in \model for each dataset by altering the hyperparameter on a base ten logarithmic scale. As shown in Figure~\ref{fig:sens}, when the hyperparameter $\lambda$ is too small or too large, the performance of the model deteriorates in most datasets, which verifies that the hyperparameter $\lambda$ is the tradeoff between the sufficiency and invariance conditions of the patterns captured by the model. 

\subsection{Complexity Analysis}

We analyze the computational complexity of \model as follows. Denote the total number of nodes and edges in the graph as $|\mathcal{V}|$ and $|\mathcal{E}|$, and the dimensionality of the hidden representation as $d$. The snapshot-wise message passing has a time complexity of $O(|\mathcal{E}|d+|\mathcal{V}|d^2)$. The fast Fourier transform has a time complexity of $O(|\mathcal{V}|d\log T)$. The disentangled spectrum mask has a time complexity of $O(|\mathcal{V}|d)$. Denote $|N_p|$ as the number of nodes or edges to predict and $S$ as the sampling number of variant patterns. Our invariant spectral filtering has a time complexity of $O(|N_p|Sd)$ in training, and does not put extra time complexity in inference. Therefore, the overall time complexity of \model is $O(|\mathcal{E}|d+|\mathcal{V}|d^2+ |\mathcal{V}|d+|\mathcal{V}|d\log T + |N_p|Sd)$. In summary, the time complexity of \model has a linear time complexity with respect to the number of nodes and edges, which is on par with the existing dynamic GNNs.

\section{Reproducibility Details}

\subsection{Training \& Evaluation}
\paragraph{Hyperparameters} Following~\cite{zhang2022dynamic}, for all methods, we adopt the hidden dimension as 32 for Aminer and 16 for other datasets. The number of layers is set to 2,
and the models are optimized with the Adam optimizer~\cite{kingma2014adam} with a learning rate 1e-2 and weight decay 5e-7. The early stopping strategy on the validation splits is adopted, with 100 epochs for Node-Synthetic datasets and 50 epochs for other datasets. For \modelnosp, we set the sampling number of variant patterns as 1000 for Collab and Yelp, and 100 for other datasets, and $\lambda$ as 1e-4,1e-3,1e-2,1e-2,1e-2 for Collab, Aminer, Yelp, Link-Synthetic, and Node-Synthetic datasets respectively. 

\paragraph{Evaluation} For link prediction tasks, we randomly sample negative links from the nodes that actually do not have links in-between, and the number of negative links is the same as the number of positive links. All the negative and positive samples for validation and testing set are kept the same for all methods. We use the inner product of the two node representations to predict links, use cross-entropy as the loss function $\ell$, and use Area under the ROC Curve (AUC) as the evaluation metric. For node classification tasks, we use a two-layer MLP for the node classifier, use cross-entropy as the loss function $\ell$, and use Accuracy (ACC) as the evaluation metric. We randomly run the experiments three times, and report the average results and standard deviations. 

\paragraph{Details of \model} 
For the node classification dataset Aminer, we conduct the missing graph trajectory complementation as follows. In practice, dynamic graphs usually encounter with the issues of incomplete trajectories, i.e., the nodes have missing historical trajectories for some reasons. For example, on academic citation networks, the papers on dynamic graphs are always different each year and they only have structures (cite other papers) at the published year, which means that they only have a one-year trajectory. In these cases, the modeling of dynamics would be difficult and also inaccurate. To complement the missing historical graph trajectories, we utilize the current structure as the virtual past structure to help model the neighborhood evolution for the node to predict at $t'$, and the message passing is 

\begin{equation}
\label{eq:gnn_miss}
\mathbf{m}^t_{u \rightarrow v} \leftarrow \mathrm{MSG}(\mathbf{h}^t_u, \mathbf{h}^t_v), 
\mathbf{h}^t_v \leftarrow\operatorname{AGG} (\{\mathbf{m}^t_{u \rightarrow v} \mid u \in \mathcal{N}^t(v) \bigcup \mathcal{N}^{t^\prime}(v) \}, \mathbf{h}^t_v\}).
\end{equation}
In this way, the node embedding $\mathbf{h}^t_u$ for node $u$ which appears at time $t\leq t^\prime $ denotes the neighborhood information it may aggregate if it appears at time $t$. Note that in Eq.~\eqref{eq:gnn_miss}, the target is to predict the node labels at time $t^\prime$, where the current neighborhood  $\mathcal{N}^{t^\prime}(v)$  is known to all methods and this method does not exploit extra future information. For the message and aggregation functions, we adopt DIDA~\cite{zhang2022dynamic} for Yelp dataset and GAT~\cite{velivckovic2017graph} for other datasets. We adopt two-layer MLPs for both the invariant and variant node classifiers.

\subsection{Dataset Details}

We summarize the dataset statistics in Table~\ref{tab:data} and describe the dataset details as follows.

\textbf{Collab}~\cite{Tang:12KDDCross,zhang2022dynamic}\footnote{https://www.aminer.cn/collaboration.} is an academic collaboration dataset with papers that were published during 1990-2006, where the nodes and edges represent author and coauthorship respectively. The author features are obtained by averaging the embeddings of the author-related papers, which are extracted by word2vec~\cite{mikolov2013efficient} from the paper abstracts. The distribution shift comes from different fields, including "Data Mining", "Database", "Medical Informatics", "Theory" and "Visualization". We use 10,1,5 chronological graph slices for training, validation and testing respectively. 

\textbf{Yelp}~\cite{sankar2020dysat,zhang2022dynamic}\footnote{https://www.yelp.com/dataset} is a business review dataset, where the nodes and edges represent customers or businesses and review behaviors respectively. We utilize the data from January 2019 to December 2020, and select users and reviews with interactions of more than 10. We use word2vec~\cite{mikolov2013efficient} to extract 32-dimensional features from the reviews and average to obtain the user and business features. The distribution shift comes from the out-break of COVID-19 midway as well as the different business categories including "Pizza", "American (New) Food", "Coffee \& Tea ",  "Sushi Bars" and "Fast Food". We use 15,1,8 chronological graph slices for training, validation and testing respectively.

\textbf{Aminer}~\cite{Tang:08KDD,sinha2015overview} is a citation network extracted from DBLP, ACM, MAG, and other sources. We select the top 20 venues, and the task is to predict the venues of the papers. We use word2vec~\cite{mikolov2013efficient} to extract 128-dimensional features from paper abstracts and average to obtain paper features. The distribution shift may come from the out-break of deep learning. We train on papers published between 2001 - 2011, validate on those published in 2012-2014, and test on those published since 2015. 

\textbf{Link-Synthetic}~\cite{zhang2022dynamic} introduces 
manual-designed distribution shift on Collab dataset. Denote the original features and structures in Collab as $\mathbf{X}_1^t$ and structures as $\mathbf{A}^t$. We introduce features $\mathbf{X}_2^t$ with a variable correlation with the labels, which are obtained by training the embeddings $\mathbf{X}_2^t \in \mathbb{R}^{N\times d}$ with the reconstruction loss  $\ell(\mathbf{X}_2^t{\mathbf{X}_2^{t}}^\top,\tilde{\mathbf{A}}^{t+1})$, where $\tilde{\mathbf{A}}^{t+1}$ refers to the sampled links, and $\ell$ refers to the cross-entropy loss function. In this way, the generated features can have strong correlations with the sampled links. 
For each time $t$, we uniformly sample $p(t)|\mathcal{E}^{t+1}|$ positive links and $(1-p(t))|\mathcal{E}^{t+1}|$ negative links in $\mathbf{A}^{t+1}$ and the sampling probability $p(t)=\text{clip}(\overline p + \sigma cos(t),0,1)$ refers to the intensity of shifts.
By controlling the parameter $p$, we can control the correlations of $\mathbf{X}^t$ and labels $\mathbf{A}^{t+1}$ to vary in training and test stage. Since the model observes the $\mathbf{X}^t = [\mathbf{X}_1^t || \mathbf{X}_2^t]$ simultaneously and the variant features are not marked, the model should discover and get rid of the variant features to handle distribution shifts. Similar to Collab dataset, we use 10,1,5 chronological graph slices for training, validation and test respectively.

\textbf{Node-Synthetic} introduces manually designed distribution shifts for node classification tasks, by simulating that some frequency components on dynamic graphs have invariant correlations with labels while some others do not. We adopt a stochastic block model (SBM)~\cite{HOLLAND1983109} to generate links between nodes. For brevity, we denote the SBM model as $\text{SBM}(\mathbf{p}_{in},p_{out})$, where $\mathbf{p}_{in} \in [0,1]^{C \times 1}$ and $p_{out}$ denotes the link probability between the nodes belonging to the same class and the link probability between the nodes from different classes respectively. We adopt $C=5$ classes. Based on the class label, each node has two types of parameters $f_{low} \in \{0.02, 0.04, 0.08, 0.10, 0.12\}$ and $f_{high} \in \{0.22, 0.24, 0.28, 0.30, 0.32 \}$. The correlation of $f_{low}$ with labels is set to 0.4, 0.6, 0.8 respectively for training and validation and 0 for testing, and the correlation of $f_{high}$ with labels is set to 1 for all data splits. The dynamic graph $\mathcal{G}^t$ at time $t$ is constructed by mixing multiple graphs together, including a random graph $\mathcal{G}^t_r$ generated from Gaussian noises, a graph constructed by the invariant parameter $\mathcal{G}^t_{I} = \text{SBM}(\mathbf{p}_{in}^{high}(t), p_{out})$ and a graph constructed by the variant parameter $\mathcal{G}^t_{I} = \text{SBM}(\mathbf{p}_{in}^{low}(t), p_{out})$. The relationship between the parameters and the link probability is  $\mathbf{p}_{in}^{low}(t, f) = S_1 (2 + \cos(2 \pi f t ))$ and $\mathbf{p}_{in}^{high}(t, f) = S_2 (2 + \cos(2 \pi f t ))$. We set 1e-3, 1e-2, 5e-3 for $p_{out}$, $S_1$ and $S_2$ respectively. We generate 4-dimensional random features for each node. On this dataset, to have better generalization ability, the model should discover and focus on the dynamic graph constructed with the invariant parameter to make predictions. 

\begin{table}[]
\caption{The summary of the dataset statistics.}
\label{tab:data}
\adjustbox{max width = \textwidth}{
\begin{tabular}{ccccccc}
\toprule
\textbf{Dataset} & \textbf{\# Snapshots} & \textbf{\# Nodes} & \textbf{\# Links} & \textbf{Time Granularity} & \textbf{\# Features} & \textbf{Evolving Features} \\ \midrule
Collab           & 16                     & 23,035            & 151,790           & Year                      & 32                  & No                         \\
Yelp             & 24                     & 13,095            & 65,375            & Month                     & 32                  & No                         \\
Link-Synthetic   & 16                     & 23,035            & 151,790           & -                         & 64                  & Yes                        \\
Aminer           & 17                     & 43,141            & 851,527           & Year                      & 128                 & No                         \\ 
Node-Synthetic   & 100                    & 5,000             & 11,252,385        & -                         & 4                   & No                        \\ \bottomrule
\end{tabular}
}

\end{table}

\subsection{Configurations}
All the experiments are conducted with:
\begin{itemize}[leftmargin=0.5cm]
    \item Operating System: Ubuntu 20.04.5 LTS
    \item CPU: Intel(R) Xeon(R) Gold 6240 CPU @ 2.60GHz
    \item GPU: NVIDIA GeForce RTX 3090 with 24 GB of memory
    \item Software: Python 3.9.12, Cuda 11.3, PyTorch~\cite{paszke2019pytorch} 1.12.1, PyTorch Geometric~\cite{Fey/Lenssen/2019} 2.0.4.
\end{itemize} 

\end{document}